%% file: acl_latex.tex
\pdfoutput=1
\PassOptionsToPackage{table}{xcolor}
\documentclass[11pt]{article}

\usepackage[preprint]{acl}

\usepackage{times}
\usepackage{latexsym}
\usepackage{times}
\usepackage{soul}
\usepackage{url}

\usepackage[flushleft]{threeparttable}
\usepackage{threeparttable}

\usepackage[utf8]{inputenc}
\usepackage{caption}
\usepackage{graphicx}
\usepackage{amsmath}
\usepackage{amsthm}
\usepackage{booktabs}
\usepackage{algorithm}
\usepackage{algorithmic}
\usepackage{pifont}
\usepackage{comment}
\usepackage{natbib}
\usepackage{wrapfig}
\usepackage{framed}
\usepackage{enumitem}
\usepackage{hyperref}
\usepackage{authblk}

\usepackage{graphicx}
\usepackage{array}
\usepackage{caption}
\usepackage{multirow}

\usepackage{multirow}
\usepackage{amsmath}  % For mathematical symbols and environments
\usepackage{amsfonts}  % For mathematical fonts (e.g., \mathbb)
\usepackage{graphicx}  % For including graphics (not used in this case)
\usepackage{bm}        % For bold symbols (e.g., \bm for bold math)
\usepackage{mathtools} % For additional math tools
\usepackage{dsfont}
\usepackage{amssymb}
\usepackage{subfigure}
\usepackage{multirow}
\usepackage{transparent}
\usepackage{amsmath,amsthm,amssymb}
\usepackage{sidecap}
\usepackage{tablefootnote}
\usepackage{authblk}
\usepackage[symbol]{footmisc}

\newtheorem{theorem}{Theorem}[section]

\newtheorem{definition}[theorem]{Definition}

\usepackage[T1]{fontenc}
\usepackage{subcaption}

\usepackage{microtype}

\usepackage{inconsolata}

\usepackage{graphicx}
\usepackage{color}
\definecolor{shadecolor}{rgb}{0.92,0.92,0.92}

\definecolor{lightgray}{gray}{0.9}
\definecolor{mediumlightgray}{gray}{0.7}
\definecolor{mediumgray}{gray}{0.5}
\definecolor{darkgray}{gray}{0.3}

\title{Save It All: Enabling Full Parameter Tuning for Federated Large Language Models via Cycle Block Gradient Descent}

\author{Lin Wang\thanks{Equal Contribution.}, Zhichao Wang$^{*}$, 
Xiaoying Tang
 \\
School of Science and Engineering, The Chinese University of Hong Kong (Shenzhen) 
}

\begin{document}
\maketitle
\begin{abstract}
The advent of large language models (LLMs) has revolutionized the deep learning paradigm, yielding impressive results across a wide array of tasks. However, the pre-training or fine-tuning of LLMs within a federated learning (FL) framework poses substantial challenges, including considerable computational and memory resource demands, as well as communication bottlenecks between servers and clients.
Existing solutions either make the unrealistic assumption that the entire model is exchanged for training, or apply parameter-effective fine-tuning methods from centralized learning to train LLMs in FL which tend to underperform during training or fine-tuning stages due to the limited search subspace of parameter updating.
In this paper, we introduce a novel method for the efficient training and fine-tuning of LLMs in FL, with minimal resource consumption. Our approach, termed FedCyBGD, utilizes Cycle Block Gradient Descent to periodically update the model. In particular, we design a compression scheme for FedCyBGD, aiming to further decrease the model download cost. It enables full parameter training in FL with only selected block updates and uploads, thereby reducing communication, computation, and memory costs. Our method achieves state-of-the-art performance for FL LLM training, while significantly reducing associated costs. Codes are provided \hyperlink{here}{https://github.com/L3030/FedCyBGD}.
\end{abstract}

\section{Introduction}

Large language models (LLMs) have demonstrated exceptional performance across a multitude of domains, including language understanding~\cite{brown2020language, devlin2018bert,achiam2023gpt,meta2024introducing,yang2024harnessing}, computer vision~\cite{radford2021learning,fang2023eva}, reasoning~\cite{zhuang2023open}, and speech recognition~\cite{radford2023robust}. By pre-training on extensive datasets, these models can learn general representations that prove beneficial for a wide array of downstream tasks~\cite{xiao2023offsite,zhuang2023efficiently}.

\begin{figure}[!t]
\centering
\includegraphics[width=.48\textwidth]{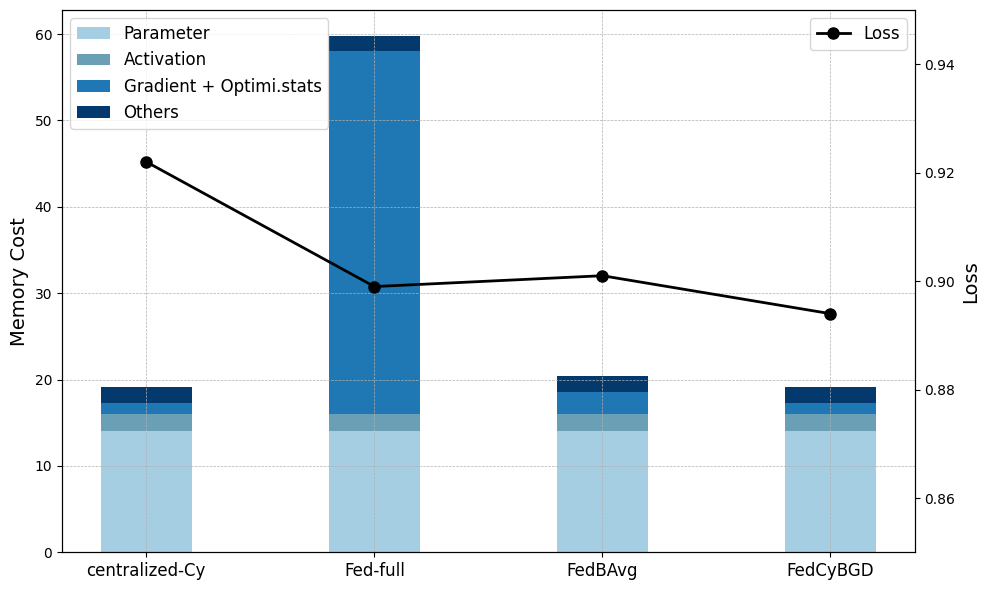}
\caption{\textbf{Observation on Federated Learning.}
Bar graphs represent the estimated memory usage for full parameter tuning of an LLaMA-7B model on a single device and the line graph represents the loss across different training paradigms. 'Centralized-Cy' denotes centralized training with cyclical block updates, 'Fed-full' refers to federated full parameter tuning with complete model communication to clients, 'FedBAvg' signifies federated training with block updates where the server selects clients for tuning and aggregates updates, and 'FedCyBGD' represents our approach, where clients cyclically participate in block tuning.
}
\label{Illuatration of observation}
\end{figure}

Despite the significant impact of LLMs in artificial intelligence and other fields, their deployment on edge devices, such as smartphones and IoT devices, is hindered by high storage and processing resource requirements. Furthermore, the growing concern over data privacy issues~\cite{yang2020federated,li2021survey,mothukuri2021survey} has led to an increased interest in federated learning (FL)~\cite{mcmahan2017communication,kairouz2019advances}, indicating a rising demand for edge-based training. 
FL is a distributed learning paradigm that enables edge devices to collaboratively train models without sharing raw data~\citep{mcmahan2017communication}. This approach addresses privacy and communication concerns and has proven valuable in sectors such as healthcare~\citep{kaissis2020secure}, energy~\citep{saputra2019energy}, and manufacturing~\citep{qu2020blockchained}. Given its data silo nature, federated learning has emerged as an effective solution for privacy-preserving LLM training~\cite{fan2023fate,kuang2023federatedscope,ye2024openfedllm}. 
Traditional FL assumes that the client possesses sufficient computational power, typically in the form of GPUs, to train computation-intensive models~\cite{wang2021comprehensive}. However, with the increasing size of model parameters and datasets, this assumption becomes increasingly impractical to fulfill in real-world scenarios~\cite{tao2022training,che2023federated}.

Notably, the pre-training or fine-tuning of LLMs incurs substantial memory and computational costs that are prohibitive for edge devices~\cite{zhao2024galore}, and communication costs remain a primary bottleneck when applied in FL~\cite{wu2024cg}. 
% \textcolor{red}{Remove this: The memory requirements encompass not only billions of trainable parameters but also their gradients and optimizer states (such as momentum and variance in Adam), which can exceed the storage of the parameters themselves~\cite{raffel2020exploring,chowdhery2023palm}. Moreover, training or fine-tuning LLMs within the FL paradigm, where edge devices have strict bandwidth limits, is particularly constrained by upload speeds~\cite{raje2024communication}. }
Some works have employed parameter-effective-fine-tuning (PEFT) for fine-tuning LLMs in FL, such as Lora, which fine-tunes the LLM based on local datasets. The principle of PEFT is to represent parameter updates in a much lower-dimensional subspace, which can potentially limit downstream performance~\cite{zhang2024scaling}. We will discuss the advantages of our framework over PEFT methods in detail in the next section and show that most existing PEFT methods are compatible with our framework.

Consequently, the challenges of training LLMs for FL and the objective of our paper can be summarized as follows:
\begin{center}
    \textit{Reducing computation, memory, and communication costs; Enabling effective federated large language model training.}
\end{center}
To address these challenges and achieve this goal, we propose a simple yet effective training paradigm called \textbf{FedCyBGD}, which trains LLMs in a \textbf{Fed}erated Learning environment using a \textbf{Cy}clic \textbf{B}lock \textbf{G}radient \textbf{D}escent approach. In terms of memory, computation, and communication costs, \textbf{FedCyBGD} aims to \textbf{\emph{save it all}}. It employs an alternating minimization strategy, training \emph{responsible blocks} on edge devices and periodically syncing their updates to the server. For \emph{unresponsible blocks}, devices download a surrogate network, a scaled-down model with similar functionality but fewer parameters, negating the need for additional updates or uploads. This method not only minimizes resource consumption during each server-client communication round but also safeguards client data privacy and ensures the preservation of the server's model privacy.
\begin{figure}[!t]
\centering
\includegraphics[width=.48\textwidth]{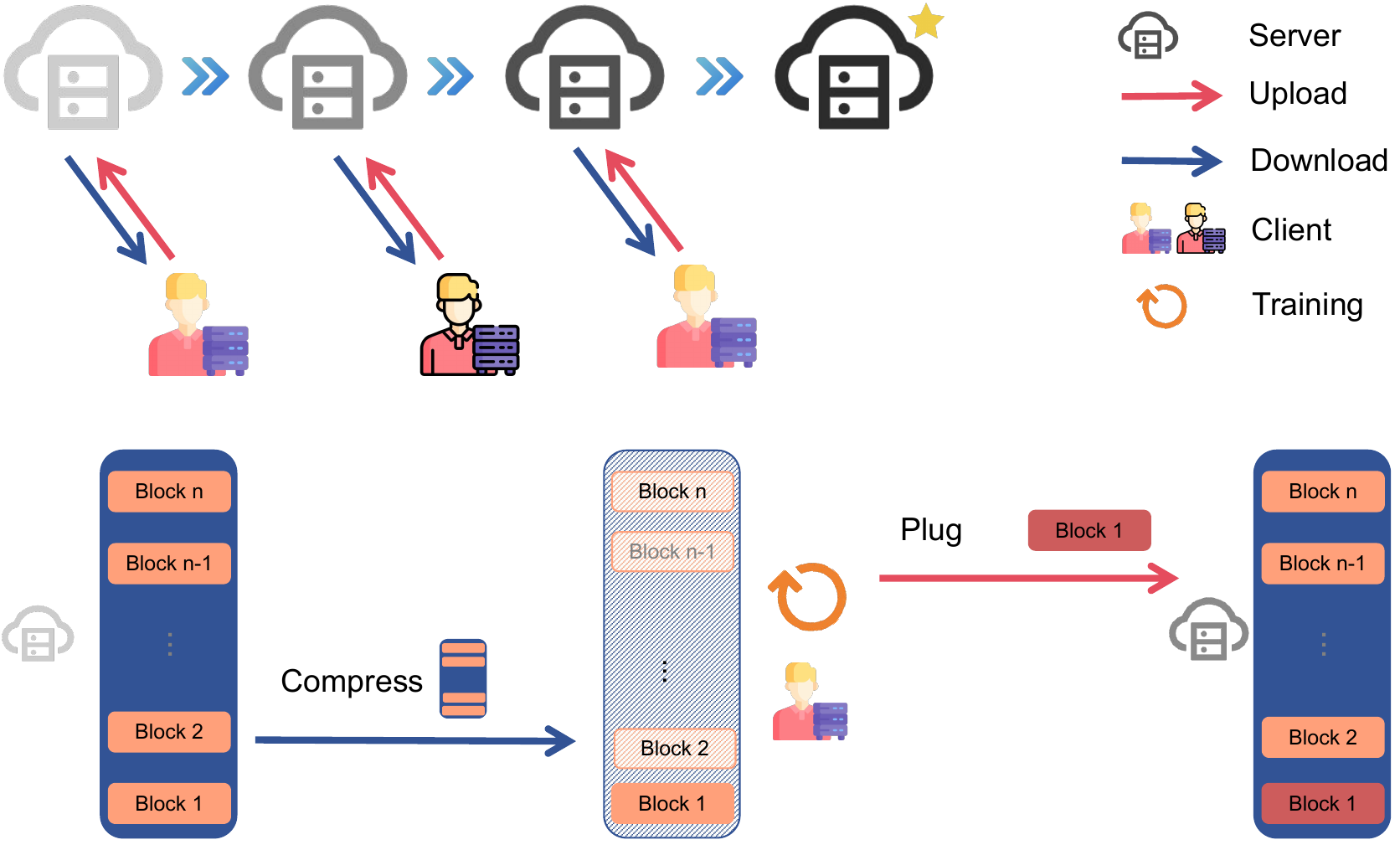}
\caption{\textbf{Overview of FedCyBGD training.} 
In FedCyBGD, the server sends responsible (Block 1) and compressed unresponsible blocks (Others) to the client. The client fine-tunes the responsible block with local data, using the frozen compressed model. The refined block is returned to the server for integration. The process, involving only compressed parameters for download and block for upload, minimizes communication parameters. By updating only responsible blocks per client, computational and memory costs are reduced, enabling full-parameter tuning for FL on resource-limited edge devices.
% Moreover, the cyclic update framework reduces the communication rounds by a factor of N compared to FedAvg. By only updating the responsible block and keeping the rest frozen, the client minimizes memory and communication costs.
}
\label{overview of FedCyGD}
\end{figure}

Extensive experiments on various public LLMs and datasets, including GPT-2~\cite{radford2019language}, RoBERTa~\cite{liu2019roberta}, ChatGLM-3~\cite{du2022glm}, BLOOM~\cite{le2023bloom}, and LLaMA-2~\cite{touvron2023LLaMA}, underscore the resource efficiency and competitive performance of our FedCyBGD framework across a variety of NLP tasks, including text generation, classification, and question answering. Our thorough evaluation assesses the framework's resource costs in terms of communication, memory, and computation. 
% Results show that Cycle Block Gradient Descent surpasses model aggregation for FL, and block updates outperform full parameter training. 
In summary, FedCyBGD provides a practical, resource-effective full parameter tuning approach for FL.

The contributions of this study can be summarized as follows:
\begin{itemize}
    \item \textbf{Cycle Block Update framework for full parameter tuning in FL}. 
    Our framework, FedCyBGD, introduces a novel FL training approach that allows LLM training in the FL paradigm by cyclically updating the LLM blocks by clients. 
    To the best of our knowledge, this is the first FL framework that enables LLM full parameter tuning on resource-limited edge devices, reducing computation, memory, and communication costs while maintaining comparable performance.
    \item \textbf{Compression for model download and block upload.}
    FedCyBGD incorporates designed compression schemes that enable the client to download only a parameter scaled-down proxy model, which reduces the client's download communication cost while keeping the responsible block properly updated.
    % applies different compression methods for various modules, applying a Power-BlockGD and surrogate module for model communication. This enables model privacy and communication reduction.
    \item 
    \textbf{Empirical Evaluation with resource Consumption Reduction.}
    % Theoretically, we analyze the computation, communication, and memory complexity of FedCyBGD, and show it is superior to other full parameter tuning methods. 
    % Empirically, 
    Extensive experiments on various public LLMs across diverse datasets demonstrate FedCyBGD's significant performance in reducing resource usage (reduced to 13.8\%) and improving training performance (up to 58.3\% higher), while preserving model and data privacy.
\end{itemize}

\section{Related Works}
\paragraph{Federated Learning for LLMs.}
The training process of Large Language Models (LLMs) can generally be divided into two stages: pre-training and fine-tuning~\cite{zhou2023comprehensive,touvron2023LLaMA,yang2024harnessing,min2023recent}. Pre-training typically involves adapting the LLM to large datasets, which usually requires comprehensive training of all parameters, a process also known as full fine-tuning. However, this approach can be prohibitively expensive, especially for federated learning where edge devices have limited capabilities~\cite{imteaj2021survey,ro2022scaling}.

Parameter Efficient Fine-Tuning (PEFT) methods such as Low-Rank Adaptation (LoRA)~\cite{hu2021lora,raje2024communication}, Adapter~\cite{houlsby2019parameter}, prompt- and prefix-tuning~\cite{li2021prefix,lester2021power}, among others, are crucial for fine-tuning LLMs under memory resource constraints. The main idea of PEFT is to represent parameter updates in a much lower-dimensional subspace, significantly reducing memory consumption. 

Despite the success of PEFT methods, fine-tuning within a substantially lower-dimensional subspace could potentially limit downstream performance~\cite{zhang2024scaling,luo2024badam}. Moreover, it requires knowledge of the entire model weights, which could compromise the privacy of data or model owners. Additionally, the fine-tuning process remains resource-intensive, as it necessitates at least one copy of the entire model to be placed on the device~\cite{xiao2023offsite}.

Several studies have built upon PEFT methods in the context of FL for LLM, including FedPETuning~\cite{zhang2023fedpetuning}, Federated Adapter Tuning~\cite{cai2022fedadapter}, Federated Prompt Tuning~\cite{zhao2023fedprompt}, and FedLora~\cite{yi2023fedlora}. However, these methods often sacrifice performance to reduce communication and computation overhead. While FL can benefit from PEFT methods, it still lacks a way to achieve full parameter tuning with an acceptable resource cost. In this work, we propose the cycle block gradient descent method, FedCyBGD, which makes it possible to save on communication, computation, and memory costs while adopting approximately full-parameter tuning. In addition, our framework is orthogonal to the PEFT methods, which can be utilized for further accelerating the tuning with reduced cost, as we showed in experiments of Section~\ref{sec experiment}.

% There are three points to clarify regarding PEFT:

% \begin{enumerate}
%     \item FedCyBGD has been shown to be effective in resource reduction compared with existing PEFT methods. Moreover, in terms of performance improvement, FedCyBGD is naturally and provably suitable for FL training, as demonstrated by our experiments.
%     \item Since FedCyBGD plays a role in directly updating the original model parameters, it is orthogonal to existing PEFT methods like LoRa. This can further reduce communication, as shown in Table~\ref{} in Section ~\ref{}.
%     \item We have designed a unique PEFT method for FedCyBGD. Specifically, we extend the communication-effective method PowerSGD~\cite{vogels2019powersgd} into Power-Block-GD to reduce the trained block. Meanwhile, we use transfer learning methods, like OffSite~\cite{xiao2023offsite}, to tune the remaining blocks. Our aim is to reduce the communication cost in each download, even though the upload is the main bottleneck for FL training.
% \end{enumerate}

\paragraph{Compression for FL LLM communication.}
Model compression techniques aim to reduce the size of a machine learning model by altering its weights or structure \cite{zhu2023survey}. Various compression methods have been developed to train computation-intensive fundamental models, including pruning \cite{han2015deep}, quantization \cite{jacob2018quantization,xiao2023smoothquant}, layer-drop \cite{sajjad2023effect}, and knowledge distillation \cite{hinton2015distilling}, all of which aim to reduce the number of parameters transferred.

In the context of LLM training in FL, various compression algorithms have proven to be effective solutions for reducing communication, such as distilling knowledge between the server and clients~\citep{tao2022training, he2020group,fan2024fedmkt}. However, these methods often rely on the server's auxiliary dataset, which is typically scenario-constrained.
Similarly, \cite{wu2024cg} compresses the communicated gradients by designing an autoencoder, while \cite{fang2024automated} designs an auto-low rank approximation method for the adapter. 
% \textcolor{red}{Inspired by the low-rank adaptation, we adapt PowerSGD~\cite{vogels2019powersgd}, a low-rank approximation method for compressing gradients, to reduce communication. This is incorporated into our trained block, termed Power-BGD, to reduce the upload cost.}
In our framework, we have minimized the main communication bottleneck, i.e., model uploading in FL~\cite{raje2024communication}. We aim to further reduce the downloading cost by avoiding the need to download the entire pre-trained model from the server. Instead, we propose using a smaller model that includes the necessary block. By utilizing this smaller model, we can update the corresponding block in the large model.
To scale down the large model while maintaining its role in local updating, we apply model pruning~\cite{ma2023llm} and layer drop~\cite{xiao2023offsite,men2024shortgpt,kim2024shortened} to the frozen blocks, reducing the downloading cost.

\section{The Proposed Method}
In this section, we introduce FedCyBGD, a novel, simple, and effective training paradigm. It enables full-parameter tuning with resource costs acceptable for edge devices by allowing each client to cyclically participate in training one or more responsible blocks. We begin by providing an intriguing observation: cyclical updates are more beneficial for LLM tuning than the traditional FL paradigm with model aggregation. We then define the specific problem addressed in this study, followed by a detailed introduction to our approach.

% In this section, we introduce FedCyBGD, an innovative and
% parameter-efficient federated mutual knowledge transfer for
% large and small language models framework. This framework
% encompasses fine-tuning scenarios in an FL setting where an
% LLM is run on the server while SLMs are distributed among
% the clients. A notable aspect of this architecture is the heterogeneity in designs among the client SLMs and their contrast
% with the server LLM. We begin by defining the specific problem addressed in this study, followed by a detailed introduction to our approach. Finally, we delve into the computational
% and communication complexities and the privacy-preserving
% aspects of our FedMKT framework.

\begin{algorithm}[!t]
\small
    \caption{\small FedCyBGD}
    \label{alg:algorithm}
    \textbf{Input}:Number of clients $m$, learning rate $\eta_l$ and $\eta_{g}$, number of local epoch $K$, total training rounds $T$, client-block partition scheme $\pi$. \\
    % \textbf{Parameter}: Optional list of parameters\\
    \textbf{Output}: Trained model parameter $\theta^T$. \\
     \textbf{Initialize:} Model parameter $\theta_0$. 
    \begin{algorithmic}[1] %[1] enables line numbers
    \FOR{$t=0$ to $T-1$}
        \STATE Assign each client $i$ a responsible block based on partition scheme $\pi = \{\pi_1, \ldots, \pi_m \}$; 
        \FOR{Client $i$ in order} % \COMMENT{Cycle participation}
            \STATE Send the compressed global model $\{\tilde{\theta}_{\pi_1}^{t+1}, \ldots, \tilde{\theta}_{\pi_{i-1}}^{t+1}, \theta_{\pi_i}^{t}, \tilde{\theta}_{\pi_{i+1}}^t, \ldots, \tilde{\theta}_{\pi_m}^t\}$ to client $i$, where $\tilde{\theta}$ is compressed model;
            \FOR{$k=0$ to $K-1$}
    \STATE $\begin{aligned}
        g_{\pi_i}^{t, k} &\leftarrow \frac{\partial}{\partial \theta_{\pi_i}} \mathcal{L}(\tilde{\theta}_{\pi_1}^{t+1}, \ldots, \tilde{\theta}_{\pi_{i-1}}^{t+1}, \theta_{\pi_i}^{t, k-1}, \\
                         &\quad \tilde{\theta}_{\pi_{i+1}}^t, \ldots, \tilde{\theta}_{\pi_m}^t)
    \end{aligned}$
\ENDFOR
            \STATE  $\Delta_t^i=\theta_{\pi_i}^{t, K-1}-\theta_{\pi_i}^{t,0}=-\eta_L\sum_{k=0}^{K-1} g_{\pi_i}^{t,k}$;
            \STATE Upload updated block ${\Delta_t^i}$;
        \ENDFOR
        \STATE Server Update Block by $\theta_{\pi_i}^{t+1} = \theta_{\pi_i}^{t} + \eta {\Delta}_t^i$;
    \ENDFOR
    \end{algorithmic}
\end{algorithm}

\subsection{Observation} 
Figure~\ref{Illuatration of observation} presents intriguing observations for FL.

The first one is that the cyclical block update paradigm outperforms the traditional FL approach involving multiple clients and aggregation. Even with less computational resource cost, the cyclical update approach easily surpasses the performance of the aggregation paradigm.
The first observation suggests that cyclical updates are an effective training paradigm for FL, outperforming the traditional aggregation strategy. This is because cyclical updates can potentially mitigate data heterogeneity and conflicting directions in aggregation. Aggregation forces different clients to focus on the same block update, amplifying the impact of heterogeneity. However, when clients update the entire model, their different datasets may lead to different update directions, reducing the likelihood of conflicts.

Another noteworthy observation is that block updates, in a cyclical manner, show superior performance compared to full parameter tuning in FL.
The second observation indicates that the block update paradigm could be more effective than full model training. It offers more granular updates and potentially alleviates the problem of catastrophic forgetting caused by frequent updates. Similar findings are reported in \citet{li2023fedtp}, which demonstrates that personalizing each client's attention layer is more effective due to negative impacts of data heterogeneity in FL.

The observations inspire us to consider applying the cycle block model update as a more effective parameter tuning paradigm for LLM tuning in FL.

\subsection{Problem Definition}

In our Federated Learning (FL) framework, a single server manages a Large Language Model (LLM) $\mathcal{M}_{\theta}$, with parameters $\theta = \{ \theta_1, \theta_2, \ldots, \theta_{|B|} \}$, where $B$ denotes the model's blocks, such as the 12 transformer blocks in GPT-2-small, thus $|B| = 12$. There are $m$ clients, each possessing a unique local dataset $\mathcal{D}_i$, with $\mathcal{D}_i \neq \mathcal{D}_j$ for $i \neq j$.

The goal of FL is to collaboratively pre-train or fine-tune the LLM using the clients' local datasets without data exchange, aiming to update the model from $\mathcal{M}_{\theta}$ to $\mathcal{M}_{\theta+\Delta}$, where $\Delta= \arg \min _\delta \mathcal{L}(\theta+\delta, \{\mathcal{D}_i\}_{i=1}^{m})$. The objective is formulated as follows:
\begin{align}
    \theta^* &= \arg \min_{\theta} \mathcal{L} ( \theta ; \mathcal{D})  \notag \\
    &= \arg \min_{\theta} \frac{1}{ m |\mathcal{D}_i|}{\sum_{i\in [m]}\sum_{\xi_i\in \mathcal{D}_i}}\mathcal{L} (\theta; \xi_i ) \, ,
\end{align}
where $\xi_i\in \mathcal{D}_i$ is the local training data of client $i$.

However, from the above equation, we can conclude that it needs full parameter training, which is uncomfortable for clients as it consumes vast communication, computation, and memory. Fortunately, training one block $\theta_{\pi_i}$ for client $i$ is more than acceptable, thus the objective can be approximated by:
\begin{align}
    \theta^* &= \{ \theta_1^*,\ldots, \theta_{|B|}^* \}  \notag\\
    &= \arg \min_{\theta_1,\ldots, \theta_{|B|}} \mathcal{L} (\theta_1,\ldots, \theta_{|B|}; \{\mathcal{D}_i\}_{i=1}^{m} ) \, .
\label{block objective}
\end{align}

Ideally, we separately update each block of the model, in order to finish the training or fine-tuning of the model within the resource-limited edge devices, and achieve similar performance compared to directly optimizing the whole model $\mathcal{M}_{\theta}$ on overall data.

% Settings. Given the foundation model $\mathcal{M}$ parameterized by $\Theta$ and the downstream dataset $\mathcal{D}$, fine-tuning the model on the downstream datasets yields $\mathcal{M}_{\Theta} \rightarrow \mathcal{M}_{\Theta+\Delta}, \Delta=$ $\arg \min _\delta \mathcal{L}(\Theta+\delta, \mathcal{D})$. To enable private and efficient transfer learning, we want to find a substitute model $\mathcal{M}_{\Theta^*}^*$ (also called as Emulator) that is (significantly) smaller and weaker than $\mathcal{M}_{\Theta}$, so that sharing $\mathcal{M}^*$ with downstream users would not threaten the ownership of the foundation models. Data owners then optimize the substitute model on the dataset, yielding $\mathcal{M}_{\Theta^*+\Delta^*}^*$. We hope that plugging the trained weights $\Delta^*$ back to the original model (i.e., $\mathcal{M}_{\ominus+\Delta^*}$ ) can achieve similar performance compared to directly optimizing $\mathcal{M}$ on the dataset (i.e., $\mathcal{M}_{\Theta+\Delta}$), without giving access to $\mathcal{M}$ itself.

\subsection{FedCyBGD}

% FedCyBGD represents an innovative and parameter-efficient federated LLM training framework. This framework enables full-parameter tuning  letting client cycle participate in training one/some responsible blocks. 

From above discussion, our primary objective is to reduce the resource cost while enabling full parameter LLM tuning for FL. We achieve this by proposing FedCyBGD, the cycle block update-based model tuning paradigm for FL, as illustrated in Figure~\ref{overview of FedCyGD} and describe the associated training algorithm in Algorithm~\ref{alg:algorithm}.

% Its primary objective it to reduce the resource cost while achieve better pre-training and fine-tuning performance in FL paradigm. We illustrate the FedCyBGD in Figure~\ref{overview of FedCyGD} and describe the associated training algorithm in Algorithm~\ref{alg:algorithm}.

\paragraph{Cycle Block Gradient Descent.}
A notable challenge that arises in objective~\ref{block objective} is to ensure that each block converges to the optimal model $\theta^*$. Inspired by Adam Block Coordinate Descent~\cite{luo2024badam}, we can approximate the update for each block using the following problem:
% \begin{equation}
% \left\{
% \begin{aligned}
\begin{align*}
    \theta_{1}^{t+1}  &= \arg\min_{\theta_1} \mathcal{L}\left(\theta_{1}, \theta_{2}^{t}, \ldots, \theta_{|B|}^t \right) \\
    \theta_{2}^{t+1}  &= \arg\min_{\theta_2} \mathcal{L}\left(\theta_{1}^{t+1}, \theta_{2}, \ldots, \theta_{|B|}^t \right) \\
   &\ ~~~~~~~~~~ \ldots \notag \\
    \theta_{|B|}^{t+1} & = \arg\min_{\theta_{|B|}} \mathcal{L}\left(\theta_{1}^{t+1}, \theta_{2}^{t+1}, \ldots, \theta_{|B|} \right) 
% \end{aligned}
% \right.
\label{Bcd update}
\end{align*}
% \end{equation}

Although the block update scheme allows for updating the large model progressively by using small blocks, it is a centralized learning paradigm that requires all data to participate in each block update. To address this challenge and leverage block coordinate descent for FL, we adopt the cycle update approach~\cite{cho2023convergence}. This approach divides clients into different groups, and these groups take turns participating in FL training.

Based on our observation that aggregation is not beneficial for training fundamental models in FL, we treat each client as a block optimizer LLM tuning. Specifically, we select one client at a time to participate in training the responsible block, while keeping the other blocks fixed at their up-to-date values.

At the $t$-th block-epoch, updating the current active block of client $i$ requires solving the following problem:
\begin{small}
\begin{align}
     \theta_{\pi_i}^{t+1} &\leftarrow {\arg \min }_{\theta_{\pi_i}} \ \mathbb{E}_{\xi_i \in \mathcal{D}_i} \notag\\
    & \left\{ \mathcal{L} \left(\theta_{\pi_1}^{t+1}, \ldots, \theta_{\pi_{i-1}}^{t+1}, \theta_{\pi_i}, \theta_{\pi_{i+1}}^t, \ldots, \theta_{\pi_m}^t ; \xi_i \right) \right\} \, ,
\end{align}
\end{small}
where $\pi_{i}$ is the block partition scheme for client $i$, and $\theta_{\pi_{i}}$ represents the responsible block for client $i$.
In this way, we utilize one client's local data to update a specific block of the LLM, alternating the parameter updates.

Since each client only trains and uploads one block, the memory, computation cost, and communication cost are significantly reduced. We will discuss the cost improvement in more detail later.

\paragraph{Compress LLM Downloading.}

The cycle block update method successfully reduces communication costs compared to traditional FL by having only one client train and upload the \emph{responsible blocks} per cycle. However, it requires clients to download the entire model to update the responsible blocks, even though other \emph{unresponsible blocks} remain frozen and aren't uploaded, leading to full model download requirements.

To tackle this challenge, we must devise a method to send only the responsible blocks to the client for updates, without transferring the entire model. This compels us to compress the model during transmission while preserving the integrity of responsible blocks. The other parts of the model, referred to as the emulator in \cite{xiao2023offsite}, role in providing approximate gradient directions for updating the responsible blocks and closely resemble the frozen original components. 
The compression process with a randomized compression operator can be formulated as follows:
% To address this issue, we need to find way to transfer the responsible block to client for updating, while not downloading the complete model. This requirement forces us to compress the transmitted model while keep the responsible unchanged. These unupdated  model, called emulator in \cite{xiao2023offsite}, used for  provide the rough gradient
% directions to update the responsible blocks while remaining similar to
% the original frozen component. The emulator in \cite{xiao2023offsite} is fixed to be the middle layer of LLM, while in our paper, we take the block besides responsible block as emulator, termed as auxiliary blocks. 

% We now introduce the notion of a randomized compression operator which is used to compress the gradients.
\begin{definition}
A randomized map $\mathcal{C}: \mathbb{R}^d \mapsto \mathbb{R}^d$ is an $\omega$-compression operator if
\begin{small}
\begin{align}
    \mathbb{E}[\mathcal{C}(\theta)]=\theta, \  \mathbb{E}\left[\|\mathcal{C}(\theta)-\theta\|^2\right] \leq \omega\|\theta\|^2, \  \forall \theta \in \mathbb{R}^d
\label{compression operator}
\end{align}
\end{small}
In particular, no compression $(\mathcal{C}(\theta) \equiv \theta)$ implies $\omega=0$.
\end{definition}
% Note that the conditions \eqref{compression operator} require the compression operator to be unbiased and its variance to be uniformly bounded by a relative magnitude of the vector being compressed.

% The compression process can be performed using mean squared error (MSE) as the loss function, as shown in the following equation:
% \begin{align}
%     \mathcal{L}_{comp}^i = \frac{1}{N_i}\sum_{i=1}^{N_i} \|\mathcal{C}(x) - x \|^2
% \end{align}
% Then, the Learnig Objective can be formulated as:
% \begin{align}
% \mathcal{L} = \mathcal{L}_1(A)
% \end{align}

There are various ways to compress the original model $\{\theta_1, \ldots, \theta_i, \ldots, \theta_{|B|}\}$ into a low-parameter model, i.e., $\mathcal{C}(\theta)_i = \{\tilde{\theta}_{\pi_1}, \ldots, \theta_{\pi_i}, \tilde{\theta}_{\pi_m} \}$, where $\tilde{\theta}$ represented the compressed model parameter.  Motivated by the experimental results from OffSite \cite{xiao2023offsite}, we use a layer drop-based compression method that provides the best balance between the aforementioned criteria. Specifically, this method randomly drops out some layers from the frozen component, the the communicated model of client $i$ can be formulated as:

\begin{align}
    \mathcal{C}(\theta)_i = \{\mathbb{I}_1\theta_1, &\ldots, \mathbb{I}_{i-1}\theta_{i-1}, \notag \\
    & \theta_i, \mathbb{I}_{i+1}\theta_{i+1}, \ldots, \mathbb{I}_{m}\theta_m\} \, ,
\end{align}
where $\mathbb{I}$
represents the indicator function, $\mathbb{I} = 0$ indicates that the layer should be dropped, which is controlled by a drop ratio.
% A larger drop ratio leads to better communication savings, while a smaller drop ratio leads to better performance.

Considering the cyclical update approach, with the current module as a reference point, the modules updated previously and those pending update exist in two distinct states, necessitating different compression strategies. For instance, if we uniformly treat all blocks, there's a risk of discarding all previously updated blocks in one round, rendering client $i$ unable to benefit from prior model updates, thus leading to ineffective tuning. To address this, we devise a hybrid pruning method: we apply low-granularity pruning to the updated blocks and layer dropping to the yet-to-be-updated blocks. Consequently, the compression operator for client $i$ in round $t$ is:
\begin{align}
    \mathcal{C}(\theta)_i^{t} =\{ \mathbb{P}_1&(\theta_1^{t+1}), \ldots, \mathbb{P}_{i-1}(\theta_{i-1}^{t+1}), \notag \\
    & \theta_i^t, \mathbb{I}_{i+1}\theta_{i+1}^{t}, \ldots, \mathbb{I}_{m}\theta_m^t \} \, ,
\end{align}
where $\mathbb{P}$ represents a generic pruning function that can be dynamically chosen according to the task. In our paper, we follow the method described by ~\citet{ma2023llm} for pruning the updated blocks. The compression process is described in Algorithm~\ref{Hybrid pruning algorithm}.

\begin{algorithm}[t]
\small
    \caption{\small Hybrid Pruning}
        \label{Hybrid pruning algorithm}
    \begin{algorithmic}[1]
        \STATE{\textbf{Input:} Server model $\theta$, Training round $T$.
        }
        \STATE{\textbf{Output:} Compressed Model $\tilde{\theta}$ to Client $i$.}
        \STATE{\textbf{Initialize:}} Round $t$, Updated Blocks Set $\mathbf{B}$, Non-updated Block Set $\overline{\mathbf{B}}$. 
        \FOR{block $\theta_j$ in $\theta = \{\theta_1, \ldots,\theta_m\}$}
            \IF{block has been updated, i.e., $\theta_j$ = $\theta_j^{t+1}$} 
                \STATE{$\mathbf{B} \gets \mathbf{B} \cup \theta_j $}
            \ELSIF{$\theta_j = \theta_j^t$} 
                \STATE{$\overline{\mathbf{B}} \gets \overline{\mathbf{B}} \cup \theta_j $}
            \ENDIF
        \ENDFOR
        \FOR{$j \in [1,\ldots, m]$}
        \STATE{Compress:
        $
        \begin{aligned}
            \mathcal{C}(\theta)_j = 
              \begin{cases}
                \mathbb{P}_j(\theta_j) & \text{if } j \in \mathbf{B}, \\
                \mathbb{I}_j \theta_j & \text{if } j \in \overline{\mathbf{B}}.
              \end{cases}
        \end{aligned}
        $
        }
        \STATE{$\tilde{\theta} \gets \mathcal{C}(\theta)_j$}
        \ENDFOR
    \end{algorithmic}
\end{algorithm}

\paragraph{Connection to PEFT.}
Our framework is orthogonal to existing parameter-efficient fine-tuning methods for LLMs, such as LoRA \cite{hu2021lora} and Adaptor \cite{houlsby2019parameter}, as shown in Section~\ref{sec experiment}. We focus on fully tuning the model with a cycle block update approach to reduce federated learning resource costs. Unlike PEFT, which uses low-rank parameter matrix approximation, our framework focuses on full parameter updates and can easily integrate PEFT by incorporating the parameter approximation into the updated blocks.

\paragraph{Computation, Memory, and Communication Complexity.} FedCyBGD offers significant computational, memory, and communication efficiency by reducing the number of parameters updated and communicated during training. For example, when using FedCyBGD to fine-tune all parameters in LLaMA-2 7B compared to traditional full tuning in FL, it reduces $25\%$ computation cost for a client for backward; each client requires about 17.3 GB, approximately 30\% of the memory cost of traditional FL full parameter tuning; and it needs only 3.5 GB for parameter communication, reducing communication cost by roughly 75\%. This results in faster training times and lower memory and communication requirements, making full parameter tuning of LLMs in an FL setting more feasible.

\section{Numerical Results}
\label{sec experiment}
In this section, we evaluate the proposed FedCyBGD on 5 datasets over 13 tasks in terms of memory consumption, running time, convergence, and downstream performance.

\begin{table*}[t]
\centering
\resizebox{.7 \textwidth}{!}{
\begin{tabular}{lccccccccccccc}
\toprule
\textbf{Method} & \textbf{Basic Model} & \textbf{Memory Cost}  & \textbf{Download Cost} & \textbf{Upload Cost}\\ \midrule
Fed-full & 12.61 GB & 62.8 GB & 12.48 GB  & 12.48 GB\\ 
FedLora & 25.24 GB & 27.4 GB & 24.96 GB & 32.03 MB  \\ 
FedAdaptor & 12.68 GB & 15.8 GB & 12.55 GB & 62.93MB \\ 
FedCyBGD-Power & 12.61 GB & 16.4 GB & 6.24 GB & 770.37 MB \\ 
FedCyBGD & 12.61 GB & 17.2 GB & 6.24 GB & 770.37 MB \\ 
\bottomrule
\end{tabular}
}
\caption{Actual memory and communication costs of applying mixed precision training to finetune LLaMA-2 7B with gradient checkpointing. Note that Fed-full only supports FP16 precision training~\tablefootnote{To mitigate potential issues like Floating Point Exceptions or loss=NaN when using pure BF16 or FP16 for LoRA fine-tuning, we opted for BF16 mixed-precision training. As a result, FedLoRA still requires storing the model in Float32 format.}. The maximum input sequence length is 4096 and the batch size is 2. }
\label{experiment memory and comm consumption}
\end{table*}

% It is worth noting that, to further save computational overhead, our experiments were conducted in pure half-precision (BF16). However, due to potential Floating Point Exceptions or loss=NaN issues that may arise when executing LoRA fine-tuning with pure BF16 or FP16, which can lead to unsuccessful training, we performed BF16 mixed-precision training on LoRA. This means that FedLoRA still needs to store the model in Float32.

\begin{table}[t]
\centering
\resizebox{.5 \textwidth}{!}{
\begin{tabular}{lccc|c} % 在倒数第二列和最后一列之间添加竖线
\toprule
\textbf{Methods} & \textbf{Forward} &  \textbf{Backward} & \textbf{Update} & \textbf{Round Time}  \\ 
\midrule
Fed-full & \ul{0.031} s & 0.449 s & 0.103 s & 20030.8 s\\ 
FedLora & 0.096 s & 0.404 s & \textbf{0.001} s & 34386.3 s\\ 
FedAdaptor & 0.069 s & 0.404 s & \ul{0.002} s & 25462.0 s\\ 
FedCyBGD-Power & 0.037 s & \textbf{0.302} s & 0.003 s & \ul{18641.1} s  \\ 
FedCyBGD & \textbf{0.030} s & \ul{0.338} s & 0.102 s & \textbf{18359.1} s\\ 
\bottomrule
\end{tabular}
}
\caption{The time spent per epoch on forward, backward, and update processes for fine-tuning the LLaMA 2-7B model on Alpaca-GPT4 is detailed. The term 'round time' refers to the time taken per global round of fine-tuning, which is completed when all clients finish their participation. FedCyBGD-Power refers to using the PowerSGD as an optimizer instead of using the ADAM optimizer. The best results are highlighted in \textbf{bold} and the second bests are marked with \ul{underline}.}
\label{Time spent}
\end{table}

\renewcommand{\arraystretch}{1.2} %控制行高
\begin{table*}[!t]
	\centering
	\begin{threeparttable}
 \resizebox{1. \textwidth}{!}{
		\begin{tabular}{ccccccccccccc}
			\toprule
			\textbf{Method}& \textbf{Memory} & \textbf{Communication}
			&\textbf{CoLA}    &\textbf{MRPC}   &\textbf{MNLI}   &\textbf{MNLI-mm}    &\textbf{QNLI}  &\textbf{QQP}   &\textbf{RTE}   &\textbf{SST-2}  &\textbf{stsb}  \cr
			\midrule
            Fed-full&  15.87GB & 915.52MB &0.57&0.87&0.88& 0.88&0.93&0.91&0.73&0.94&0.90 \\
		  FedLora & 13.67GB & 463.36MB &0.45&0.72&0.54&0.55&0.68&0.75&0.55&0.82&0.59\\
		FedAdaptor &  13.12GB & 466.97MB &0.42&0.71&0.54&0.55&0.71&0.76&0.56&0.82&0.43\\
			\hline
		FedCyBGD &  \cellcolor{mediumlightgray}2.79GB & \cellcolor{lightgray}591.93MB
                & 0.52 & 0.80 & 0.87 & 0.87 & 0.92 & 0.90 & 0.57 & 0.93 & 0.88 \\
			\bottomrule
		\end{tabular}
  }
	\end{threeparttable}
 \caption{Evaluating FedCyBGD for full parameter tuning on GLUE benchmark using pre-trained RoBERTa-base. Higher scores on these tasks represent better performance.}
\label{tab glue}
\end{table*}
% \renewcommand{\arraystretch}{1.2} %控制行高
% \begin{table*}[tp]
% 	\centering
% 	\begin{threeparttable}
%  \resizebox{1. \textwidth}{!}{
% 		\begin{tabular}{ccccccccccccc}
% 			\toprule
% 			Method& Comp Time $[Forward, Backward, Step]$& Memory & Comm
% 			&CoLA    &Mrpc   &mnli   &mnli-mm    &qnli  &qqp   &rte   &sst2  &stsb  \cr
% 			\midrule
%             Fed-FT& $[0.006, 0.008, 0.003]$ 
% 			& PC & |&0.57&0.87&0.88& 0.88&0.93&0.91&0.73&0.94&0.90 \\
% 		  FedLora & $[0.011, 0.008, 0.001]$& SC & | &0.45&0.72&0.54&0.55&0.68&0.75&0.55&0.82&0.59\\
% 		FedAdaptor & $[0.012, 0.007, 0.001]$ & |RE& | &0.42&0.71&0.54&0.55&0.71&0.76&0.56&0.82&0.43\\
% 			\hline
% 		FedCyBGD & \cellcolor{$[0.013, 0.006, 0.008]$} & \cellcolor{mediumlightgray} Cell & \cellcolor{mediumgray} Cell
%                 & 0.52 & 0.80 & 0.87 & 0.87 & 0.92 & 0.90 & 0.57 & 0.93 & 0.88 \\
% 			\bottomrule
% 		\end{tabular}
%   }
% 	\end{threeparttable}
%  \caption{Evaluating FedCyBGD for full parameter tuning on GLUE benchmark using pre-trained RoBERTa-base. We report the average score of all tasks.}
% \label{tab:perfor}
% \end{table*}
\begin{table*}[h!]
\small
\centering
	\begin{threeparttable}
 \resizebox{.75\textwidth}{!}{
 \small
\begin{tabular}{llccccc}
\toprule
\textbf{Task} & \textbf{Method} & \textbf{ChatGLM-6B} & \textbf{Bloom-1.3B} & \textbf{LLaMA2-7B} \\ \midrule
\multirow{6}{*}{\textbf{Alpaca-GPT4} $\downarrow$} 
& Fed-full & 1.27 & \textbf{1.47} & 0.92   \\ 
 & BADAM & \ul{1.22} & 1.57 &  0.90  \\ 
 & FedLora & - & 1.64 &  \ul{0.90}  \\ 
 & FedAdaptor & 1.26 & 1.57 & 0.90\\
  & FedCyBGD-Power & 1.26 &1.93 & 0.93\\
 & FedCyBGD & \textbf{1.21} & \ul{1.56} & \textbf{0.90}\\
 \midrule
\multirow{6}{*}{\textbf{OASST1} $\downarrow$}
& Fed-full & 1.85 & \textbf{1.85} & 1.14   \\ 
 & BADAM & 1.69 & 1.88 &  1.14  \\ 
 & FedLora & - & 1.92 &  1.14  \\ 
 & FedAdaptor & \ul{1.72}& 1.90 & \ul{1.12}\\
  & FedCyBGD-Power & 2.13 & 2.09& 1.13\\
 & FedCyBGD & \textbf{1.62} & \ul{1.86}& \textbf{1.12}\\
 \bottomrule
\end{tabular}
}
\end{threeparttable}
\caption{Performance Comparison of FedCyBGD and baseline approaches. We assessed FedCyBGD's generalization across three models and two tasks. Identical values with different ranks (\textbf{bold} for best, \ul{underline} for second-best) indicate unseen decimal differences.}
\label{alpaca task}
\end{table*}

\begin{table}[h!]
\centering
 \resizebox{.5 \textwidth}{!}{
\begin{tabular}{lcccc}
\hline
            & FedFT    & BADAM & FedAdapter  & FedCyBGD \\ \toprule
Exact Match $\uparrow$ & \textbf{68.14} & 47.32 & 9.07 & \ul{67.36}   \\
F1      $\uparrow$    & \textbf{79.03}  & 60.29 & 16.28 & \ul{77.82}   \\
\hline
Memory   &  16.80GB & 15.21GB & 13.65GB & 15.21GB   \\
Communication         & 974MB  & 514MB & 493.8MB & 514MB   \\
\bottomrule
\end{tabular}
}
\caption{ Comparison of FedCyBGD and baselines on SQuAD QA task. All results are obtained using GPT-2 small, batch size 24. Exact Match and F1 are two evaluation metrics for QA task}
\label{squad task}
\end{table}

\begin{table}[h!]
\centering
\resizebox{.5 \textwidth}{!}{
\begin{tabular}{lccccc}
\hline
Method & Performance $\downarrow$ & Memory Cost & Communication Cost\\ \toprule
FedCyBGD  & 0.9    &  17.20GB  & 13.23GB \\
\quad + i  & 0.91  &  17.20GB & 6.99GB \\
\quad + ii & 1.04  & 26.89GB &  12.49GB\\
\quad + iii  & 1.11    &  14.20GB & 6.30GB \\ \bottomrule
\end{tabular}
}
\caption{The performance of FedCyBGD, compatible with compression and PEFT methods, on Alpaca-GPT4 dataset with LLaMa-2. Here, FedCyBGD refers to the cyclic block update without compressed download. i) represents the hybrid compression method; ii) represents the LoRA method with rank 16; and iii) refers to the Adaptor tuning method.}
\label{ablation on CyBGD}
\end{table}

\subsection{Experimental Setup}
%减小client number来加速实验，甚至只用3个

\paragraph{Models And Datasets.} We evaluate FedCyBGD on large language models, including GPT2-small~\cite{radford2019language}, BLOOM~\citep{le2023bloom}, RoBERTa-base~\citep{liu2019roberta}, ChatGLM3-6B~\cite{du2022glm} and LLaMA2-7B~\cite{touvron2023LLaMA}. We evaluate language models on the GLUE benchmark~\cite{wang2018glue}, which is a benchmark including 10 widely used NLP datasets for evaluating the performance on a variety of tasks, including sentiment analysis, question answering, and textual entailment; and four other tasks including instruction following task on Alpaca-GPT4~\cite{peng2023instruction}, OpenAssistant Conversations task on OASST1~\cite{kopf2024openassistant}, and  QA task on SQuAD~\citep{rajpurkar2016squad}. For FL environment, we consider resource-limited 32 and 64 devices, and a parameter server. In each cycle, randomly choose a cyclic order for updating. 

\paragraph{Baselines:}
We conduct a comparative analysis of our FedCyBGD framework against several baselines. These include LLM-ZS, which represents the zero-shot capabilities of LLM; Fed-full, a full parameter tuning method; two parameter-effective fine-tuning methods, namely Adaptor~\cite{houlsby2019parameter} and Lora~\cite{hu2021lora}; BADAM~\cite{luo2024badam}, another centralized effective LLM training method; and two optimization methods, ADAM~\cite{kingma2014adam} and PowerSGD~\cite{vogels2019powersgd}. Unless otherwise specified, we use the ADAM optimizer as the backbone optimization method. For a fair comparison, we adapt the centralized PEFT methods, specifically Adapter and Lora, to the federated learning setting.
More details about the experiment setup are provided in Appendix~\ref{app sec exp}.
% Experiment description refers to the FATE-LLM~\cite{fan2023fate}, 

% We compare with FL-LLM baselines for demonstrating that our frameworks is a practical pipeline for \textbf{(1)training} and \textbf{finetuning} LLM for FL. 

% Model: LLaMA2-7B, gpt-2-mid

% Dataset: xx

% Baseline:

% \begin{itemize}
%     \item LLM-ZS, which represent the zero-shot capabilities of LLM on the server; 
%     \item LLM-FT , which represents the full parameter tuning in the centralized manner
%     \item full parameter cycle tuning (BADAM, cycle centralized); 
%     \item (fix and random) aggregation of FL tuning;
%     \item (fix and random) cycle training;
%     \item \emph{FedLora},
%     \item \emph{FedAdaptor}
% \end{itemize}

% \renewcommand{\arraystretch}{1.5} %控制行高
% \begin{table*}[tp]
% 	\centering
% 	\fontsize{7.5}{10}\selectfont
% 	\begin{threeparttable}
% 		\caption{Actual memory and communication costs of applying mixed precision training to finetune LLaMA 3-8B with gradient checkpointing using a single RTX3090.}
% 		\label{memory and communication consumption }
% 		\begin{tabular}{cccccccccccccc}
% 			\toprule
% 			\multirow{1}{*}{Method}&  Parameter 
% 			&Gradient &Optim.states&Memory consump &Communication Comsumption\cr
% 			\midrule
% 			% \multirow{4}{*} 
%             Full Tuning & PC &0.48&0.22&0.01&0.42\cr
% 			BADAM & SC &0.76&0.90&0.04&0.48\cr
% 			LoRA-rank & |RE& 8.43&8.09&8.13&8.78\cr
%             FedCyBGD & |RE& 8.43&8.09&8.13&8.78\cr
% 			\bottomrule
% 		\end{tabular}
% 	\end{threeparttable}
% \end{table*}

\subsection{Evaluation of FedCyBGD}
\paragraph{Memory and Communication Consumption.}
Table~\ref{experiment memory and comm consumption} demonstrates that FedCyBGD is effective in reducing the communicated parameter and memory costs compared to existing full-parameter tuning paradigms or the PEFT training method. Specifically, it reduces memory consumption by approximately 72.7\% compared to existing full-parameter tuning methods. In terms of communication, even when compared to the best-performing PEFT method, our full-parameter tuning framework reduces communication costs by about 44.1\%.

\paragraph{Wall-clock running time comparison.}
The time consumption for each method is divided into forward, backward, and update phases, as detailed in Table~\ref{Time spent}. For the LLaMA2 model fine-tuning, the forward times of Fed-full and FedCyBGD are similar, as both process the entire model. FedLora and FedAdaptor incur slightly more time due to extra steps in activation registration and low-rank adapter computations, respectively. Notably, FedCyBGD cuts the backward time nearly in half compared to Fed-full and FedLora.
% The time consumption of each method primarily consists of three components, i.e., forward, backward, and update. The result for fine-tuning LLaMA2 model are shown in Table~\ref{Time spent}. The forward time for full parameter-tuning method, i.e., Fed-full and FedCyBGD is rather close since they all need forward the whole model.
% The slight higher time cost for FedLora and FedAdaptor arrtibutes to additional operation for registering actibations and the calculation of the low-rank adapters, respectively. 
% Regarding backward time, FedCyBGD reduces the time cost by nearly half compared with Fed-full and FedLora.

\paragraph{Performance improvement by FedCyBGD.}
Table~\ref{tab glue} illustrates that our method achieves competitive performance with traditional full parameter tuning while significantly reducing memory and communication costs (reduced to 13.8\%). Furthermore, we demonstrate the consistent effectiveness of FedCyBGD by evaluating it on various LLMs and datasets. Specifically, in Table~\ref{alpaca task}, FedCyBGD consistently outperforms baselines in two different datasets, Alpaca-GPT4 and OASST1, using large-sized models LLaMA2-7B and ChatGLM-6B, and achieves competitive performance with full tuning in Bloom-1.3B. Additionally, in the question-answering task, FedCyBGD exhibits a significant improvement over FedAdaptor, with up to 58.3\% higher performance and minimal communication cost.

\paragraph{Orthogonal to PEFT.}
Table~\ref{ablation on CyBGD} demonstrates that FedCyBGD is compatible with existing PEFT algorithms such as LoRA and Adaptor. By utilizing PEFT, FedCyBGD successfully reduces communication and memory costs, albeit with a slight sacrifice in performance. Furthermore, the table highlights the effectiveness of our compression method, which significantly reduces communication while maintaining a relatively small impact on performance.

% \begin{table}{r}{0.5\textwidth}
% \small
% \centering
% \vspace{-.em}
% \caption{ 
% \textbf{Ablation study for FedCyBGD:} + OffSite; + PEFT.
% } 
% \resizebox{.5\textwidth}{!}{%
%   \begin{tabular}{l l l l l l l l l c c}
%    \toprule
%    \multirow{2}{*}{Algorithm} & \multicolumn{2}{c}{CIFAR-10} & \multicolumn{2}{c}{CIFAR-100}\\
%    \cmidrule(lr){2-3} \cmidrule(lr){4-5} 
%                     & $\alpha=0.1$ & $c=5$     &  $\alpha=0.1$ & $c=10$\\
%    \midrule
% FedAvg-FT  & 59.47{\transparent{0.5}±0.09} & 73.46{\transparent{0.5}±0.36} & 32.21{\transparent{0.5}±0.52} & 64.23{\transparent{0.5}±0.73} \\ 
% \quad + \emph{i} & 65.57{\transparent{0.5}±0.12} & 76.96{\transparent{0.5}±0.53} & 57.25{\transparent{0.5}±0.99}  &68.10{\transparent{0.5}±1.47}  \\ 
% \quad + \emph{i} + \emph{ii} & 68.89{\transparent{0.5}±0.08} & 78.95{\transparent{0.5}±0.20} & 58.02{\transparent{0.5}±0.83}  &69.60{\transparent{0.5}±0.91}  \\ 
% \quad + \emph{i} + \emph{ii} + \emph{iii} &71.89{\transparent{0.5}±0.06} & 80.36{\transparent{0.5}±0.32} & 62.11{\transparent{0.5}±1.82}  &71.07{\transparent{0.5}±1.02} \\ 
% \bottomrule
% \end{tabular}
% }
% \label{table of ablation of algorithm}
% \vspace{- .em}
% \end{table}

% \paragraph{Ablation on overall client number}

% Using bar and line figure

% \paragraph{Ablation on compression method}
% In our proposed compression method, the compress ratio have a significant influence on performance.

\paragraph{Additional Ablation Studies.}
To thoroughly validate the effectiveness of our method, we conducted experiments under various settings, which included different numbers of clients, diverse batch sizes, and block allocation strategies for clients, as detailed in Appendix~\ref{app sec exp}. It is noteworthy that: 1) our method demonstrates increasingly apparent memory efficiency advantages over the PEFT method as the batch size grows; and 2) our method exhibits robustness to variations in both the block allocation strategy and the number of clients.

% Table 3
% \begin{table}[h]
% \centering
% \begin{tabular}{|l|c|c|c|}
% \hline
% \textbf{Method} & \textbf{Forward} & \textbf{Backward} & \textbf{Update} \\ \hline
% LOMO & 0.78 hours & 3.70 hours & 136 seconds \\ \hline
% LoRA & 0.83 hours & 3.20 hours & 146 seconds \\ \hline
% BAdam & 0.71 hours & 1.74 hours & 142 seconds \\ \hline
% \end{tabular}
% \caption{Time spent per epoch on forward, backward, and update for finetuning LLaMA 3.8B using a single RTX3090. The single pass batch size is 2. The results are averaged over 3 epochs.}
% \end{table}

% % Table 4
% \begin{table}[h]
% \centering
% \begin{tabular}{|l|c|}
% \hline
% \textbf{Backward scheme} & \textbf{Backward time} \\ \hline
% All modules & 0.636 seconds \\ \hline
% Input module only & 0.327 seconds \\ \hline
% Output module only & 0.629 seconds \\ \hline
% \end{tabular}
% \caption{Time spent on different backward schemes with batch size 2 for finetuning LLaMA 3.8B using a single RTX3090. The results are averaged over 100 backward passes.}
% \end{table}

\section{Conclusions}
In this work, we introduce FedCyBGD, a novel, simple, and effective full parameter tuning framework for federated Large Language Models (LLMs). This pioneering approach allows users to cyclically update blocks, thereby facilitating the tuning of federated LLMs. FedCyBGD significantly reduces computation, memory, and communication costs for resource-limited edge clients in Federated Learning (FL), enabling full parameter training and tuning of LLaMA-2 7B using a single RTX 3090-24GB GPU. To the best of our knowledge, FedCyBGD is the first framework that enables full parameter LLM pre-training or fine-tuning for federated learning settings on a single consumer-level GPU. Extensive empirical results on various LLMs and NLP tasks demonstrate the framework's effectiveness in terms of accuracy (up to 58.3\% higher) and efficiency (reduced to 13.8\%).

\section{Limitations}
While our method can significantly reduce the resource cost for federated learning to achieve full parameter tuning LLMs, we still lack theoretical guarantees regarding the convergence of the cycle block update. Previous work on cycle updates analysis, such as \cite{cho2023convergence}, can be utilized in the block update paradigm. Furthermore, our focus has primarily been on applying FedCyBGD to NLP tasks, but extending its application to other tasks, such as visual tasks, presents an opportunity to showcase the capabilities of FedCyBGD. We leave these directions for future improvements.

\clearpage
\newpage
\bibliography{custom}

\clearpage
\newpage
\onecolumn
\input{Appendix}

\end{document}

%% file: Appendix.tex
\clearpage
\section{Experiment Details}
\label{app sec exp}

\subsection{Experimental Environment} 
\label{exp env}
For all experiments, we use NVIDIA GeForce RTX 3090 and H20 GPUs.

\subsection{Experimental Setup}

\paragraph{General setup.}
In this paper, we present the evaluation of FedCyBGD on several large language models, including GPT2-small~\cite{radford2019language}, BLOOM~\citep{le2023bloom}, RoBERTa-base~\citep{liu2019roberta}, ChatGLM3-6B~\cite{du2022glm}, and LLaMA2-7B~\cite{touvron2023LLaMA}. The evaluation is conducted using the GLUE benchmark~\cite{wang2018glue}, which comprises 10 widely used NLP datasets designed to assess performance across a variety of tasks such as sentiment analysis, question answering, and textual entailment. Additionally, we include four other tasks: the instruction following task on Alpaca-GPT4~\cite{peng2023instruction}, the OpenAssistant Conversations task on OASST1~\cite{kopf2024openassistant}, and the QA task on SQuAD~\citep{rajpurkar2016squad}.

For the federated learning (FL) environment, we consider scenarios involving resource-limited devices, specifically configurations with 32 and 64 devices, along with a parameter server. In each training cycle, a random cyclic order is chosen for updating the models. This setup aims to simulate realistic conditions and evaluate the robustness and efficiency of FedCyBGD under varying resource constraints.

In this section, we conduct a comparative analysis of our FedCyBGD framework against several baseline methods. The baselines include LLM-ZS, which represents the zero-shot capabilities of large language models (LLMs); Fed-full, a full parameter tuning method; and two parameter-efficient fine-tuning methods, namely Adaptor~\cite{houlsby2019parameter} and Lora~\cite{hu2021lora}. Additionally, we compare against BADAM~\cite{luo2024badam}, a centralized effective LLM training method, and two optimization methods, ADAM~\cite{kingma2014adam} and PowerSGD~\cite{vogels2019powersgd}. Unless otherwise specified, the ADAM optimizer is used as the backbone optimization method.
% \begin{table*}[ht]
% \centering
% \resizebox{1. \textwidth}{!}{
% \begin{tabular}{cccccccccccccc}
% \toprule
% \textbf{Method} & \textbf{Parameter} & \textbf{Memory consump.}  & \textbf{Download Para.} & \textbf{Upload Para.}\\ \midrule
% Fed-full & 14GB & 62.8GB & 7GB  & 7GB\\ 
% FedLora & \textcolor{red}{14}GB & 27.4GB & 7GB & 8.4M  \\ 
% FedAdaptor & \textcolor{red}{14.6}GB & 15.8GB & 7GB & 33M \\ 
% FedCyBGD-Power & 14.6GB & 16.4GB & 3.5GB & 202M \\ 
% FedCyBGD & 14GB & 17.2GB & 3.5GB & 202M \\ 
% \bottomrule
% \end{tabular}
% }
% \caption{Actual memory and communication costs of applying mixed precision training to finetune Chatglm and BLOOM with gradient checkpointing. Note that LOMO only supports FP16 precision training. The maximum input sequence length is 4096 and the batch size is 2.}
% \label{esssssion}
% \end{table*}

\begin{table}[h]
    \centering
    \begin{tabular}{lcccccc}
        \toprule
        & LLM-ZS & ChatGLM & BLOOM & GPT-2 & ReBERTa \\
        \midrule
        \textbf{batch size} & 2 & 2 & 2 & 24& 32\\
        \textbf{accumulation step} & 8 & 8 & 8 & 1 & 1  \\
        \textbf{learning rate} & 1.0e-6 & 1.0e-6& 1.0e-6 & 3.0e-5& 2.0e-5 \\
        \bottomrule
    \end{tabular}
    \caption{Training details of FedCyBGD and baselines in different models and tasks.}
    \label{tab:comparison}
\end{table}

\subsection{Additional results}
\begin{table}[h]
\centering
\resizebox{0.5 \textwidth}{!}{
\begin{tabular}{lccccccccccccc}
\toprule
\textbf{Method} &  \textbf{ChatGLM} & \textbf{BLOOM} &\textbf{LLaMA-2}  \\ \midrule
Fed-full & 47341.2 s & 29074.8 s & 20030.8 s\\ 
FedLora & - & 22815.0 s  & 34386.3 s\\ 
FedAdaptor  & - & 21261.2 s & 25462.0 s\\ 
FedCyBGD-Power & 41374.21 s & 19845.7 s& 18641.1 s  \\ 
FedCyBGD  & 40241.45 s & 18092.7s  & 18359.1 s\\ 
\bottomrule
\end{tabular}
}
\caption{Time spent per federated client participation round for fine-tuning LLaMA2-7B on Alpaca-GPT4 with batch size 2. }
\label{app Time spent}
\end{table}

\begin{figure}[h]
    \centering
    \includegraphics[width=0.5\textwidth]{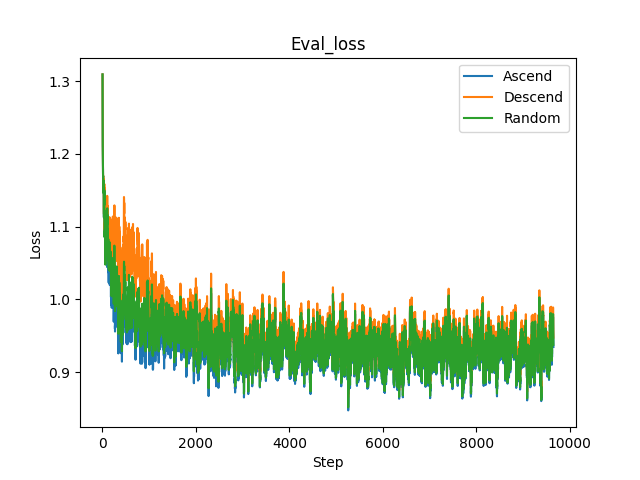}
    \caption{The performance of FedCyBGD under different block allocation strategies.}
    \label{app block allocation}
\end{figure}

% \begin{figure}
%     \centering
%     \includegraphics[width=0.5\textwidth]{Fig/eval_loss.png}
%     \caption{The performance of FedCyBGD under different block allocation strategies.}
%     \label{app block allocation}
% \end{figure}

\begin{table}[h]
\centering
\resizebox{0.3 \textwidth}{!}{
\begin{tabular}{lccccccccccccc}
\toprule
\textbf{Client Num} &  \textbf{ 32} & \textbf{64}   \\ \midrule
Fed-full & 0.94 & 0.95\\ 
FedCyBGD & 0.92 & 0.93  \\ 
\bottomrule
\end{tabular}
}
\caption{The performance under client number. In particular, the model is LLaMA-2 7B, dataset is Alpaca-GPT4. We report the result in one communication round. }
\label{app Time spesssnt}
\end{table}

\begin{table}[h]
\centering
\resizebox{0.3 \textwidth}{!}{
\begin{tabular}{lccccccccccccc}
\toprule
\textbf{batch size} &  \textbf{ 32} & \textbf{64}   \\ \midrule
Fed-full & 915.52 MB & 11513 MB\\ 
FedLora & 463.36 MB  & 8271 MB\\ 
FedLora & 466.97 MB & 8071 MB\\ 
FedCyBGD & 591.93 MB & 4805 MB  \\ 
\bottomrule
\end{tabular}
}
\caption{The memory consumption of FedCyBGD and PEFT methods with different batch sizes on MRPC dataset with RoBERTa. }
\label{app Time spesssnt}
\end{table}

% \begin{figure}[h]
%     \centering
%     \begin{subfigure}[b]{width = 0.4\textwidth}
%         \centering
%         \includegraphics[width=\textwidth]{Fig/training_eval_loss0.png}
%         \caption{Convergence of Client 1}
%         \label{fig:image1}
%     \end{subfigure}
%     \hfill
%     \begin{subfigure}[b]{width = 0.4\textwidth}
%         \centering
%         \includegraphics[width=\textwidth]{Fig/training_eval_loss.png}
%         \caption{Convergence of Client 32}
%         \label{fig:image2}
%     \end{subfigure}
%     \caption{Overall convergence performance in FedCyBGD, from beginning to ending.}
%     \label{fig:two_images}
% \end{figure}

    \begin{figure}[!ht]
     \centering
     \vspace{-0.em}
     \subfigure[Client 1\label{fairness mnist}]{ \includegraphics[width=.45\textwidth,]{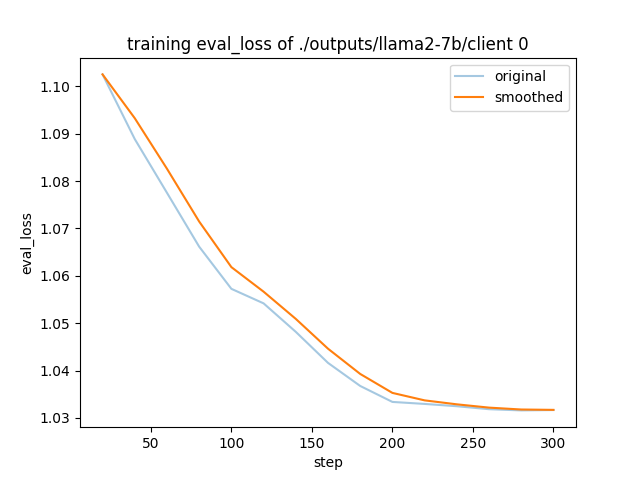}}
     % \quad
     \subfigure[Client 32\label{fairness cifar10}]{ \includegraphics[width=.45\textwidth,]{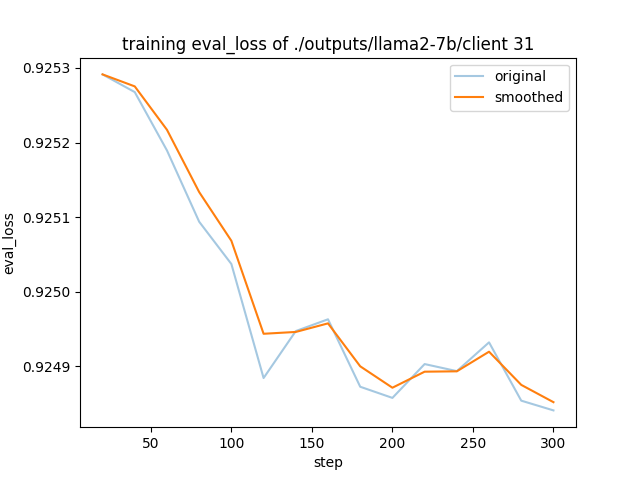}}
     \caption{Overall convergence performance in FedCyBGD, from beginning to ending.
     }
     \label{figure variance over cifar}
     \vspace{-.5em}
    \end{figure}

%% file: acl_latex.bbl
\begin{thebibliography}{67}
\providecommand{\natexlab}[1]{#1}

\bibitem[{Achiam et~al.(2023)Achiam, Adler, Agarwal, Ahmad, Akkaya, Aleman, Almeida, Altenschmidt, Altman, Anadkat et~al.}]{achiam2023gpt}
Josh Achiam, Steven Adler, Sandhini Agarwal, Lama Ahmad, Ilge Akkaya, Florencia~Leoni Aleman, Diogo Almeida, Janko Altenschmidt, Sam Altman, Shyamal Anadkat, et~al. 2023.
\newblock Gpt-4 technical report.
\newblock \emph{arXiv preprint arXiv:2303.08774}.

\bibitem[{Brown et~al.(2020)Brown, Mann, Ryder, Subbiah, Kaplan, Dhariwal, Neelakantan, Shyam, Sastry, Askell et~al.}]{brown2020language}
Tom Brown, Benjamin Mann, Nick Ryder, Melanie Subbiah, Jared~D Kaplan, Prafulla Dhariwal, Arvind Neelakantan, Pranav Shyam, Girish Sastry, Amanda Askell, et~al. 2020.
\newblock Language models are few-shot learners.
\newblock \emph{Advances in neural information processing systems}, 33:1877--1901.

\bibitem[{Cai et~al.(2022)Cai, Wu, Wang, Lin, and Xu}]{cai2022fedadapter}
Dongqi Cai, Yaozong Wu, Shangguang Wang, Felix~Xiaozhu Lin, and Mengwei Xu. 2022.
\newblock Fedadapter: Efficient federated learning for modern nlp.
\newblock \emph{arXiv preprint arXiv:2205.10162}.

\bibitem[{Che et~al.(2023)Che, Liu, Zhou, Ren, Zhou, Sheng, Dai, and Dou}]{che2023federated}
Tianshi Che, Ji~Liu, Yang Zhou, Jiaxiang Ren, Jiwen Zhou, Victor~S Sheng, Huaiyu Dai, and Dejing Dou. 2023.
\newblock Federated learning of large language models with parameter-efficient prompt tuning and adaptive optimization.
\newblock \emph{arXiv preprint arXiv:2310.15080}.

\bibitem[{Cho et~al.(2023)Cho, Sharma, Joshi, Xu, Kale, and Zhang}]{cho2023convergence}
Yae~Jee Cho, Pranay Sharma, Gauri Joshi, Zheng Xu, Satyen Kale, and Tong Zhang. 2023.
\newblock On the convergence of federated averaging with cyclic client participation.
\newblock In \emph{International Conference on Machine Learning}, pages 5677--5721. PMLR.

\bibitem[{Devlin et~al.(2018)Devlin, Chang, Lee, and Toutanova}]{devlin2018bert}
Jacob Devlin, Ming-Wei Chang, Kenton Lee, and Kristina Toutanova. 2018.
\newblock Bert: Pre-training of deep bidirectional transformers for language understanding.
\newblock \emph{arXiv preprint arXiv:1810.04805}.

\bibitem[{Du et~al.(2022)Du, Qian, Liu, Ding, Qiu, Yang, and Tang}]{du2022glm}
Zhengxiao Du, Yujie Qian, Xiao Liu, Ming Ding, Jiezhong Qiu, Zhilin Yang, and Jie Tang. 2022.
\newblock Glm: General language model pretraining with autoregressive blank infilling.
\newblock In \emph{Proceedings of the 60th Annual Meeting of the Association for Computational Linguistics (Volume 1: Long Papers)}, pages 320--335.

\bibitem[{Fan et~al.(2023)Fan, Kang, Ma, Chen, Wei, Fan, and Yang}]{fan2023fate}
Tao Fan, Yan Kang, Guoqiang Ma, Weijing Chen, Wenbin Wei, Lixin Fan, and Qiang Yang. 2023.
\newblock Fate-llm: A industrial grade federated learning framework for large language models.
\newblock \emph{arXiv preprint arXiv:2310.10049}.

\bibitem[{Fan et~al.(2024)Fan, Ma, Kang, Gu, Fan, and Yang}]{fan2024fedmkt}
Tao Fan, Guoqiang Ma, Yan Kang, Hanlin Gu, Lixin Fan, and Qiang Yang. 2024.
\newblock Fedmkt: Federated mutual knowledge transfer for large and small language models.
\newblock \emph{arXiv preprint arXiv:2406.02224}.

\bibitem[{Fang et~al.(2023)Fang, Wang, Xie, Sun, Wu, Wang, Huang, Wang, and Cao}]{fang2023eva}
Yuxin Fang, Wen Wang, Binhui Xie, Quan Sun, Ledell Wu, Xinggang Wang, Tiejun Huang, Xinlong Wang, and Yue Cao. 2023.
\newblock Eva: Exploring the limits of masked visual representation learning at scale.
\newblock In \emph{Proceedings of the IEEE/CVF Conference on Computer Vision and Pattern Recognition}, pages 19358--19369.

\bibitem[{Fang et~al.(2024)Fang, Lin, Chen, Chen, Gao, and Fang}]{fang2024automated}
Zihan Fang, Zheng Lin, Zhe Chen, Xianhao Chen, Yue Gao, and Yuguang Fang. 2024.
\newblock Automated federated pipeline for parameter-efficient fine-tuning of large language models.
\newblock \emph{arXiv preprint arXiv:2404.06448}.

\bibitem[{Han et~al.(2015)Han, Mao, and Dally}]{han2015deep}
Song Han, Huizi Mao, and William~J Dally. 2015.
\newblock Deep compression: Compressing deep neural networks with pruning, trained quantization and huffman coding.
\newblock \emph{arXiv preprint arXiv:1510.00149}.

\bibitem[{He et~al.(2020)He, Annavaram, and Avestimehr}]{he2020group}
Chaoyang He, Murali Annavaram, and Salman Avestimehr. 2020.
\newblock Group knowledge transfer: Federated learning of large cnns at the edge.
\newblock \emph{Advances in Neural Information Processing Systems}, 33:14068--14080.

\bibitem[{Hinton et~al.(2015)Hinton, Vinyals, and Dean}]{hinton2015distilling}
Geoffrey Hinton, Oriol Vinyals, and Jeff Dean. 2015.
\newblock Distilling the knowledge in a neural network.
\newblock \emph{arXiv preprint arXiv:1503.02531}.

\bibitem[{Houlsby et~al.(2019)Houlsby, Giurgiu, Jastrzebski, Morrone, De~Laroussilhe, Gesmundo, Attariyan, and Gelly}]{houlsby2019parameter}
Neil Houlsby, Andrei Giurgiu, Stanislaw Jastrzebski, Bruna Morrone, Quentin De~Laroussilhe, Andrea Gesmundo, Mona Attariyan, and Sylvain Gelly. 2019.
\newblock Parameter-efficient transfer learning for nlp.
\newblock In \emph{International conference on machine learning}, pages 2790--2799. PMLR.

\bibitem[{Hu et~al.(2021)Hu, Shen, Wallis, Allen-Zhu, Li, Wang, Wang, and Chen}]{hu2021lora}
Edward~J Hu, Yelong Shen, Phillip Wallis, Zeyuan Allen-Zhu, Yuanzhi Li, Shean Wang, Lu~Wang, and Weizhu Chen. 2021.
\newblock Lora: Low-rank adaptation of large language models.
\newblock \emph{arXiv preprint arXiv:2106.09685}.

\bibitem[{Imteaj et~al.(2021)Imteaj, Thakker, Wang, Li, and Amini}]{imteaj2021survey}
Ahmed Imteaj, Urmish Thakker, Shiqiang Wang, Jian Li, and M~Hadi Amini. 2021.
\newblock A survey on federated learning for resource-constrained iot devices.
\newblock \emph{IEEE Internet of Things Journal}, 9(1):1--24.

\bibitem[{Jacob et~al.(2018)Jacob, Kligys, Chen, Zhu, Tang, Howard, Adam, and Kalenichenko}]{jacob2018quantization}
Benoit Jacob, Skirmantas Kligys, Bo~Chen, Menglong Zhu, Matthew Tang, Andrew Howard, Hartwig Adam, and Dmitry Kalenichenko. 2018.
\newblock Quantization and training of neural networks for efficient integer-arithmetic-only inference.
\newblock In \emph{Proceedings of the IEEE conference on computer vision and pattern recognition}, pages 2704--2713.

\bibitem[{Kairouz et~al.(2019)Kairouz, McMahan, Avent, Bellet, Bennis, Bhagoji, Bonawitz, Charles, Cormode, Cummings et~al.}]{kairouz2019advances}
Peter Kairouz, H~Brendan McMahan, Brendan Avent, Aur{\'e}lien Bellet, Mehdi Bennis, Arjun~Nitin Bhagoji, Kallista Bonawitz, Zachary Charles, Graham Cormode, Rachel Cummings, et~al. 2019.
\newblock Advances and open problems in federated learning.
\newblock \emph{arXiv preprint arXiv:1912.04977}.

\bibitem[{Kaissis et~al.(2020)Kaissis, Makowski, R{\"u}ckert, and Braren}]{kaissis2020secure}
Georgios~A Kaissis, Marcus~R Makowski, Daniel R{\"u}ckert, and Rickmer~F Braren. 2020.
\newblock Secure, privacy-preserving and federated machine learning in medical imaging.
\newblock \emph{Nature Machine Intelligence}, 2(6):305--311.

\bibitem[{Kim et~al.(2024)Kim, Kim, Kim, Castells, Choi, Shin, and Song}]{kim2024shortened}
Bo-Kyeong Kim, Geonmin Kim, Tae-Ho Kim, Thibault Castells, Shinkook Choi, Junho Shin, and Hyoung-Kyu Song. 2024.
\newblock Shortened llama: A simple depth pruning for large language models.
\newblock \emph{arXiv preprint arXiv:2402.02834}.

\bibitem[{Kingma and Ba(2014)}]{kingma2014adam}
Diederik~P Kingma and Jimmy Ba. 2014.
\newblock Adam: A method for stochastic optimization.
\newblock \emph{arXiv preprint arXiv:1412.6980}.

\bibitem[{K{\"o}pf et~al.(2024)K{\"o}pf, Kilcher, von R{\"u}tte, Anagnostidis, Tam, Stevens, Barhoum, Nguyen, Stanley, Nagyfi et~al.}]{kopf2024openassistant}
Andreas K{\"o}pf, Yannic Kilcher, Dimitri von R{\"u}tte, Sotiris Anagnostidis, Zhi~Rui Tam, Keith Stevens, Abdullah Barhoum, Duc Nguyen, Oliver Stanley, Rich{\'a}rd Nagyfi, et~al. 2024.
\newblock Openassistant conversations-democratizing large language model alignment.
\newblock \emph{Advances in Neural Information Processing Systems}, 36.

\bibitem[{Kuang et~al.(2023)Kuang, Qian, Li, Chen, Gao, Pan, Xie, Li, Ding, and Zhou}]{kuang2023federatedscope}
Weirui Kuang, Bingchen Qian, Zitao Li, Daoyuan Chen, Dawei Gao, Xuchen Pan, Yuexiang Xie, Yaliang Li, Bolin Ding, and Jingren Zhou. 2023.
\newblock Federatedscope-llm: A comprehensive package for fine-tuning large language models in federated learning.
\newblock \emph{arXiv preprint arXiv:2309.00363}.

\bibitem[{Le~Scao et~al.(2023)Le~Scao, Fan, Akiki, Pavlick, Ili{\'c}, Hesslow, Castagn{\'e}, Luccioni, Yvon, Gall{\'e} et~al.}]{le2023bloom}
Teven Le~Scao, Angela Fan, Christopher Akiki, Ellie Pavlick, Suzana Ili{\'c}, Daniel Hesslow, Roman Castagn{\'e}, Alexandra~Sasha Luccioni, Fran{\c{c}}ois Yvon, Matthias Gall{\'e}, et~al. 2023.
\newblock Bloom: A 176b-parameter open-access multilingual language model.

\bibitem[{Lester et~al.(2021)Lester, Al-Rfou, and Constant}]{lester2021power}
Brian Lester, Rami Al-Rfou, and Noah Constant. 2021.
\newblock The power of scale for parameter-efficient prompt tuning.
\newblock \emph{arXiv preprint arXiv:2104.08691}.

\bibitem[{Li et~al.(2023)Li, Cai, Wang, Tang, Ding, Lin, and Shi}]{li2023fedtp}
Hongxia Li, Zhongyi Cai, Jingya Wang, Jiangnan Tang, Weiping Ding, Chin-Teng Lin, and Ye~Shi. 2023.
\newblock Fedtp: Federated learning by transformer personalization.
\newblock \emph{IEEE transactions on neural networks and learning systems}.

\bibitem[{Li et~al.(2021)Li, Wen, Wu, Hu, Wang, Li, Liu, and He}]{li2021survey}
Qinbin Li, Zeyi Wen, Zhaomin Wu, Sixu Hu, Naibo Wang, Yuan Li, Xu~Liu, and Bingsheng He. 2021.
\newblock A survey on federated learning systems: Vision, hype and reality for data privacy and protection.
\newblock \emph{IEEE Transactions on Knowledge and Data Engineering}, 35(4):3347--3366.

\bibitem[{Li and Liang(2021)}]{li2021prefix}
Xiang~Lisa Li and Percy Liang. 2021.
\newblock Prefix-tuning: Optimizing continuous prompts for generation.
\newblock \emph{arXiv preprint arXiv:2101.00190}.

\bibitem[{Liu et~al.(2019)Liu, Ott, Goyal, Du, Joshi, Chen, Levy, Lewis, Zettlemoyer, and Stoyanov}]{liu2019roberta}
Yinhan Liu, Myle Ott, Naman Goyal, Jingfei Du, Mandar Joshi, Danqi Chen, Omer Levy, Mike Lewis, Luke Zettlemoyer, and Veselin Stoyanov. 2019.
\newblock Roberta: A robustly optimized bert pretraining approach.
\newblock \emph{arXiv preprint arXiv:1907.11692}.

\bibitem[{Luo et~al.(2024)Luo, Yu, and Li}]{luo2024badam}
Qijun Luo, Hengxu Yu, and Xiao Li. 2024.
\newblock Badam: A memory efficient full parameter training method for large language models.
\newblock \emph{arXiv preprint arXiv:2404.02827}.

\bibitem[{Ma et~al.(2023)Ma, Fang, and Wang}]{ma2023llm}
Xinyin Ma, Gongfan Fang, and Xinchao Wang. 2023.
\newblock Llm-pruner: On the structural pruning of large language models.
\newblock \emph{Advances in neural information processing systems}, 36:21702--21720.

\bibitem[{McMahan et~al.(2017)McMahan, Moore, Ramage, Hampson, and y~Arcas}]{mcmahan2017communication}
Brendan McMahan, Eider Moore, Daniel Ramage, Seth Hampson, and Blaise~Aguera y~Arcas. 2017.
\newblock Communication-efficient learning of deep networks from decentralized data.
\newblock In \emph{Artificial intelligence and statistics}, pages 1273--1282. PMLR.

\bibitem[{Men et~al.(2024)Men, Xu, Zhang, Wang, Lin, Lu, Han, and Chen}]{men2024shortgpt}
Xin Men, Mingyu Xu, Qingyu Zhang, Bingning Wang, Hongyu Lin, Yaojie Lu, Xianpei Han, and Weipeng Chen. 2024.
\newblock Shortgpt: Layers in large language models are more redundant than you expect.
\newblock \emph{arXiv preprint arXiv:2403.03853}.

\bibitem[{Meta(2024)}]{meta2024introducing}
AI~Meta. 2024.
\newblock Introducing meta llama 3: The most capable openly available llm to date.
\newblock \emph{Meta AI.}

\bibitem[{Min et~al.(2023)Min, Ross, Sulem, Veyseh, Nguyen, Sainz, Agirre, Heintz, and Roth}]{min2023recent}
Bonan Min, Hayley Ross, Elior Sulem, Amir Pouran~Ben Veyseh, Thien~Huu Nguyen, Oscar Sainz, Eneko Agirre, Ilana Heintz, and Dan Roth. 2023.
\newblock Recent advances in natural language processing via large pre-trained language models: A survey.
\newblock \emph{ACM Computing Surveys}, 56(2):1--40.

\bibitem[{Mothukuri et~al.(2021)Mothukuri, Parizi, Pouriyeh, Huang, Dehghantanha, and Srivastava}]{mothukuri2021survey}
Viraaji Mothukuri, Reza~M Parizi, Seyedamin Pouriyeh, Yan Huang, Ali Dehghantanha, and Gautam Srivastava. 2021.
\newblock A survey on security and privacy of federated learning.
\newblock \emph{Future Generation Computer Systems}, 115:619--640.

\bibitem[{Peng et~al.(2023)Peng, Li, He, Galley, and Gao}]{peng2023instruction}
Baolin Peng, Chunyuan Li, Pengcheng He, Michel Galley, and Jianfeng Gao. 2023.
\newblock Instruction tuning with gpt-4.
\newblock \emph{arXiv preprint arXiv:2304.03277}.

\bibitem[{Qu et~al.(2020)Qu, Pokhrel, Garg, Gao, and Xiang}]{qu2020blockchained}
Youyang Qu, Shiva~Raj Pokhrel, Sahil Garg, Longxiang Gao, and Yong Xiang. 2020.
\newblock A blockchained federated learning framework for cognitive computing in industry 4.0 networks.
\newblock \emph{IEEE Transactions on Industrial Informatics}, 17(4):2964--2973.

\bibitem[{Radford et~al.(2021)Radford, Kim, Hallacy, Ramesh, Goh, Agarwal, Sastry, Askell, Mishkin, Clark et~al.}]{radford2021learning}
Alec Radford, Jong~Wook Kim, Chris Hallacy, Aditya Ramesh, Gabriel Goh, Sandhini Agarwal, Girish Sastry, Amanda Askell, Pamela Mishkin, Jack Clark, et~al. 2021.
\newblock Learning transferable visual models from natural language supervision.
\newblock In \emph{International conference on machine learning}, pages 8748--8763. PMLR.

\bibitem[{Radford et~al.(2023)Radford, Kim, Xu, Brockman, McLeavey, and Sutskever}]{radford2023robust}
Alec Radford, Jong~Wook Kim, Tao Xu, Greg Brockman, Christine McLeavey, and Ilya Sutskever. 2023.
\newblock Robust speech recognition via large-scale weak supervision.
\newblock In \emph{International Conference on Machine Learning}, pages 28492--28518. PMLR.

\bibitem[{Radford et~al.(2019)Radford, Wu, Child, Luan, Amodei, Sutskever et~al.}]{radford2019language}
Alec Radford, Jeffrey Wu, Rewon Child, David Luan, Dario Amodei, Ilya Sutskever, et~al. 2019.
\newblock Language models are unsupervised multitask learners.
\newblock \emph{OpenAI blog}, 1(8):9.

\bibitem[{Raje(2024)}]{raje2024communication}
Arian Raje. 2024.
\newblock \emph{Communication-Efficient LLM Training for Federated Learning}.
\newblock Ph.D. thesis, Carnegie Mellon University Pittsburgh, PA.

\bibitem[{Rajpurkar et~al.(2016)Rajpurkar, Zhang, Lopyrev, and Liang}]{rajpurkar2016squad}
Pranav Rajpurkar, Jian Zhang, Konstantin Lopyrev, and Percy Liang. 2016.
\newblock Squad: 100,000+ questions for machine comprehension of text.
\newblock \emph{arXiv preprint arXiv:1606.05250}.

\bibitem[{Ro et~al.(2022)Ro, Breiner, McConnaughey, Chen, Suresh, Kumar, and Mathews}]{ro2022scaling}
Jae~Hun Ro, Theresa Breiner, Lara McConnaughey, Mingqing Chen, Ananda~Theertha Suresh, Shankar Kumar, and Rajiv Mathews. 2022.
\newblock Scaling language model size in cross-device federated learning.
\newblock \emph{arXiv preprint arXiv:2204.09715}.

\bibitem[{Sajjad et~al.(2023)Sajjad, Dalvi, Durrani, and Nakov}]{sajjad2023effect}
Hassan Sajjad, Fahim Dalvi, Nadir Durrani, and Preslav Nakov. 2023.
\newblock On the effect of dropping layers of pre-trained transformer models.
\newblock \emph{Computer Speech \& Language}, 77:101429.

\bibitem[{Saputra et~al.(2019)Saputra, Hoang, Nguyen, Dutkiewicz, Mueck, and Srikanteswara}]{saputra2019energy}
Yuris~Mulya Saputra, Dinh~Thai Hoang, Diep~N Nguyen, Eryk Dutkiewicz, Markus~Dominik Mueck, and Srikathyayani Srikanteswara. 2019.
\newblock Energy demand prediction with federated learning for electric vehicle networks.
\newblock In \emph{2019 IEEE global communications conference (GLOBECOM)}, pages 1--6. IEEE.

\bibitem[{Tao et~al.(2022)Tao, Gao, and Guo}]{tao2022training}
Jiang Tao, Zhen Gao, and Zhaohui Guo. 2022.
\newblock Training vision transformers in federated learning with limited edge-device resources.
\newblock \emph{Electronics}, 11(17):2638.

\bibitem[{Touvron et~al.(2023)Touvron, Martin, Stone, Albert, Almahairi, Babaei, Bashlykov, Batra, Bhargava, Bhosale et~al.}]{touvron2023LLaMA}
Hugo Touvron, Louis Martin, Kevin Stone, Peter Albert, Amjad Almahairi, Yasmine Babaei, Nikolay Bashlykov, Soumya Batra, Prajjwal Bhargava, Shruti Bhosale, et~al. 2023.
\newblock Llama 2: Open foundation and fine-tuned chat models.
\newblock \emph{arXiv preprint arXiv:2307.09288}.

\bibitem[{Vogels et~al.(2019)Vogels, Karimireddy, and Jaggi}]{vogels2019powersgd}
Thijs Vogels, Sai~Praneeth Karimireddy, and Martin Jaggi. 2019.
\newblock Powersgd: Practical low-rank gradient compression for distributed optimization.
\newblock \emph{Advances in Neural Information Processing Systems}, 32.

\bibitem[{Wang et~al.(2018)Wang, Singh, Michael, Hill, Levy, and Bowman}]{wang2018glue}
Alex Wang, Amanpreet Singh, Julian Michael, Felix Hill, Omer Levy, and Samuel~R Bowman. 2018.
\newblock Glue: A multi-task benchmark and analysis platform for natural language understanding.
\newblock \emph{arXiv preprint arXiv:1804.07461}.

\bibitem[{Wang et~al.(2021)Wang, Qu, Zhou, Zhang, Luo, Xu, Guo, and Li}]{wang2021comprehensive}
Haozhao Wang, Zhihao Qu, Qihua Zhou, Haobo Zhang, Boyuan Luo, Wenchao Xu, Song Guo, and Ruixuan Li. 2021.
\newblock A comprehensive survey on training acceleration for large machine learning models in iot.
\newblock \emph{IEEE Internet of Things Journal}, 9(2):939--963.

\bibitem[{Wu et~al.(2024)Wu, Li, Zhang, Xu, Wu, Zhao, and Liu}]{wu2024cg}
Huiwen Wu, Xiaohan Li, Deyi Zhang, Xiaogang Xu, Jiafei Wu, Puning Zhao, and Zhe Liu. 2024.
\newblock Cg-fedllm: How to compress gradients in federated fune-tuning for large language models.
\newblock \emph{arXiv preprint arXiv:2405.13746}.

\bibitem[{Xiao et~al.(2023{\natexlab{a}})Xiao, Lin, and Han}]{xiao2023offsite}
Guangxuan Xiao, Ji~Lin, and Song Han. 2023{\natexlab{a}}.
\newblock Offsite-tuning: Transfer learning without full model.
\newblock \emph{arXiv preprint arXiv:2302.04870}.

\bibitem[{Xiao et~al.(2023{\natexlab{b}})Xiao, Lin, Seznec, Wu, Demouth, and Han}]{xiao2023smoothquant}
Guangxuan Xiao, Ji~Lin, Mickael Seznec, Hao Wu, Julien Demouth, and Song Han. 2023{\natexlab{b}}.
\newblock Smoothquant: Accurate and efficient post-training quantization for large language models.
\newblock In \emph{International Conference on Machine Learning}, pages 38087--38099. PMLR.

\bibitem[{Yang et~al.(2024)Yang, Jin, Tang, Han, Feng, Jiang, Zhong, Yin, and Hu}]{yang2024harnessing}
Jingfeng Yang, Hongye Jin, Ruixiang Tang, Xiaotian Han, Qizhang Feng, Haoming Jiang, Shaochen Zhong, Bing Yin, and Xia Hu. 2024.
\newblock Harnessing the power of llms in practice: A survey on chatgpt and beyond.
\newblock \emph{ACM Transactions on Knowledge Discovery from Data}, 18(6):1--32.

\bibitem[{Yang et~al.(2020)Yang, Fan, and Yu}]{yang2020federated}
Qiang Yang, Lixin Fan, and Han Yu. 2020.
\newblock \emph{Federated learning: Privacy and incentive}, volume 12500.
\newblock Springer Nature.

\bibitem[{Ye et~al.(2024)Ye, Wang, Chai, Li, Li, Xu, Du, Wang, and Chen}]{ye2024openfedllm}
Rui Ye, Wenhao Wang, Jingyi Chai, Dihan Li, Zexi Li, Yinda Xu, Yaxin Du, Yanfeng Wang, and Siheng Chen. 2024.
\newblock Openfedllm: Training large language models on decentralized private data via federated learning.
\newblock \emph{arXiv preprint arXiv:2402.06954}.

\bibitem[{Yi et~al.(2023)Yi, Yu, Wang, and Liu}]{yi2023fedlora}
Liping Yi, Han Yu, Gang Wang, and Xiaoguang Liu. 2023.
\newblock Fedlora: Model-heterogeneous personalized federated learning with lora tuning.
\newblock \emph{arXiv preprint arXiv:2310.13283}.

\bibitem[{Zhang et~al.(2024)Zhang, Liu, Cherry, and Firat}]{zhang2024scaling}
Biao Zhang, Zhongtao Liu, Colin Cherry, and Orhan Firat. 2024.
\newblock When scaling meets llm finetuning: The effect of data, model and finetuning method.
\newblock \emph{arXiv preprint arXiv:2402.17193}.

\bibitem[{Zhang et~al.(2023)Zhang, Yang, Dai, Wang, Yu, Qu, and Xu}]{zhang2023fedpetuning}
Zhuo Zhang, Yuanhang Yang, Yong Dai, Qifan Wang, Yue Yu, Lizhen Qu, and Zenglin Xu. 2023.
\newblock Fedpetuning: When federated learning meets the parameter-efficient tuning methods of pre-trained language models.
\newblock In \emph{Annual Meeting of the Association of Computational Linguistics 2023}, pages 9963--9977. Association for Computational Linguistics (ACL).

\bibitem[{Zhao et~al.(2023)Zhao, Du, Li, Li, and Liu}]{zhao2023fedprompt}
Haodong Zhao, Wei Du, Fangqi Li, Peixuan Li, and Gongshen Liu. 2023.
\newblock Fedprompt: Communication-efficient and privacy-preserving prompt tuning in federated learning.
\newblock In \emph{ICASSP 2023-2023 IEEE International Conference on Acoustics, Speech and Signal Processing (ICASSP)}, pages 1--5. IEEE.

\bibitem[{Zhao et~al.(2024)Zhao, Zhang, Chen, Wang, Anandkumar, and Tian}]{zhao2024galore}
Jiawei Zhao, Zhenyu Zhang, Beidi Chen, Zhangyang Wang, Anima Anandkumar, and Yuandong Tian. 2024.
\newblock Galore: Memory-efficient llm training by gradient low-rank projection.
\newblock \emph{arXiv preprint arXiv:2403.03507}.

\bibitem[{Zhou et~al.(2023)Zhou, Li, Li, Yu, Liu, Wang, Zhang, Ji, Yan, He et~al.}]{zhou2023comprehensive}
Ce~Zhou, Qian Li, Chen Li, Jun Yu, Yixin Liu, Guangjing Wang, Kai Zhang, Cheng Ji, Qiben Yan, Lifang He, et~al. 2023.
\newblock A comprehensive survey on pretrained foundation models: A history from bert to chatgpt.
\newblock \emph{arXiv preprint arXiv:2302.09419}.

\bibitem[{Zhu et~al.(2023)Zhu, Li, Liu, Ma, and Wang}]{zhu2023survey}
Xunyu Zhu, Jian Li, Yong Liu, Can Ma, and Weiping Wang. 2023.
\newblock A survey on model compression for large language models.
\newblock \emph{arXiv preprint arXiv:2308.07633}.

\bibitem[{Zhuang et~al.(2023{\natexlab{a}})Zhuang, Liu, Koopman, and Zuccon}]{zhuang2023open}
Shengyao Zhuang, Bing Liu, Bevan Koopman, and Guido Zuccon. 2023{\natexlab{a}}.
\newblock Open-source large language models are strong zero-shot query likelihood models for document ranking.
\newblock \emph{arXiv preprint arXiv:2310.13243}.

\bibitem[{Zhuang et~al.(2023{\natexlab{b}})Zhuang, Liu, Ning, Huang, Lv, Huang, Zhao, Zhang, Mao, Wang et~al.}]{zhuang2023efficiently}
Yan Zhuang, Qi~Liu, Yuting Ning, Weizhe Huang, Rui Lv, Zhenya Huang, Guanhao Zhao, Zheng Zhang, Qingyang Mao, Shijin Wang, et~al. 2023{\natexlab{b}}.
\newblock Efficiently measuring the cognitive ability of llms: An adaptive testing perspective.
\newblock \emph{arXiv preprint arXiv:2306.10512}.

\end{thebibliography}
